\documentclass{article}

\usepackage{arxiv}

\usepackage[utf8]{inputenc} % allow utf-8 input
\usepackage[T1]{fontenc}    % use 8-bit T1 fonts
\usepackage{hyperref}       % hyperlinks
\usepackage{url}            % simple URL typesetting
\usepackage{booktabs}       % professional-quality tables
\usepackage{amsfonts}       % blackboard math symbols
\usepackage{nicefrac}       % compact symbols for 1/2, etc.
\usepackage{microtype}      % microtypography
\usepackage{lipsum}		% Can be removed after putting your text content
\usepackage{graphicx}
\usepackage{natbib}
\usepackage{doi}
\usepackage[most]{tcolorbox}
\usepackage{multirow}   % multirow cells
\usepackage{booktabs}   % nicer horizontal rules (\toprule ...)
\usepackage{enumitem}

\title{Illusions in Humans and AI: How Visual Perception Aligns and Diverges}

\author{
    Jianyi Yang\textsuperscript{*}, 
    Junyi Ye\textsuperscript{*}, 
    Ankan Dash,
    Guiling Wang\textsuperscript{†}
}

\date{August 2025}

\hypersetup{
pdftitle={A template for the arxiv style},
pdfsubject={q-bio.NC, q-bio.QM},
pdfauthor={David S.~Hippocampus, Elias D.~Striatum},
pdfkeywords={First keyword, Second keyword, More},
}

\begin{document}
\maketitle

\begingroup
\renewcommand\thefootnote{\relax}
\footnotetext{\textsuperscript{*} Equal contribution \quad \textsuperscript{†} Corresponding author}
\endgroup

\begin{abstract}
By comparing biological and artificial perception through the lens of illusions, we highlight critical differences in how each system constructs visual reality. Understanding these divergences can inform the development of more robust, interpretable, and human-aligned artificial intelligence (AI) vision systems.
In particular, visual illusions expose how human perception is based on contextual assumptions rather than raw sensory data. As artificial vision systems increasingly perform human-like tasks, it is important to ask: does AI experience illusions, too? Does it have unique illusions?
This article explores how AI responds to classic visual illusions that involve color, size, shape, and motion. We find that some illusion-like effects can emerge in these models, either through targeted training or as by-products of pattern recognition. In contrast, we also identify illusions unique to AI, such as pixel-level sensitivity and hallucinations, that lack human counterparts. By systematically comparing human and AI responses to visual illusions, we uncover alignment gaps and AI-specific perceptual vulnerabilities invisible to human perception. These findings provide insights for future research on vision systems that preserve human-beneficial perceptual biases while avoiding distortions that undermine trust and safety.
\end{abstract}

% keywords can be removed

\section{Introduction}
\label{sec:introduction}

From phantom colors to impossible shapes, visual illusions expose a key truth: What we see is not simply what’s there, but what our brain believes should be there.
Human vision is remarkably effective, and remarkably deceptive. We effortlessly interpret the world around us: identifying objects, estimating depth, and making snap decisions. But under the special conditions, our eyes betray us.

\begin{figure}[ht]
    \centering
    \includegraphics[width=0.3\linewidth]{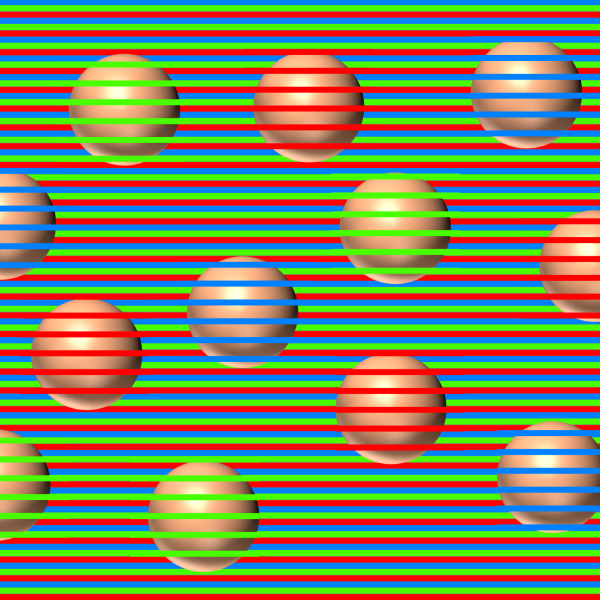}
    \includegraphics[width=0.3\linewidth]{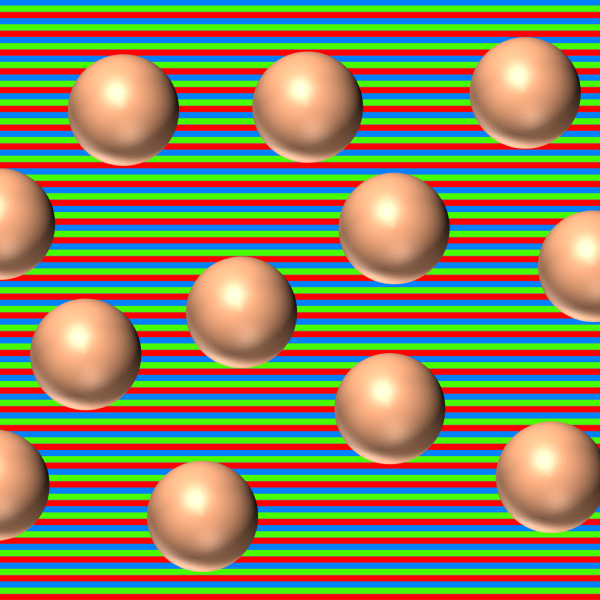}
    \includegraphics[width=0.3\linewidth]{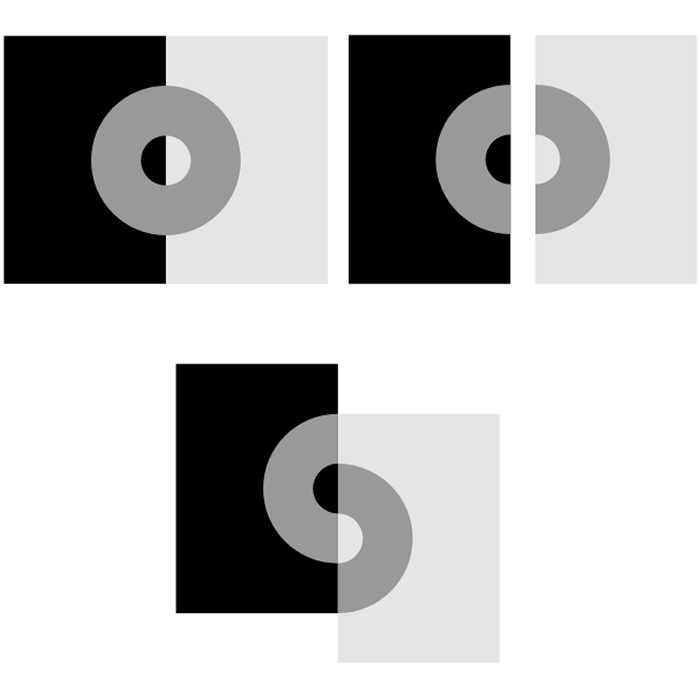}
    \caption{Color \& Brightness Illusion: Munker–White Illusion (left), the same figure without the overlaid stripes (center), and Brightness Contrast Illusion (right). Left and center images by David Novick, University of Texas at El Paso. Right image by Akiyoshi Kitaoka, Ritsumeikan University.}
    \label{fig:color}
\end{figure}

\begin{tcolorbox}[colback=gray!10, colframe=gray!40!black, boxrule=0.5pt, arc=4pt, left=6pt, right=6pt, top=4pt, bottom=4pt]
To see this effect in action, look at the left image in Figure~\ref{fig:color}. What color do the spheres appear to be? Now examine the center image, where the colored stripes have been removed from the spheres: do they still appear to be different colors?
\end{tcolorbox}

This is the \textbf{Munker–White illusion} \cite{Munker–White_Illusion}, a powerful example of how our visual system integrates context to make sense of ambiguous input. It challenges the idea that perception is a passive reflection of the world; instead, it is an active, inferential process.

As AI systems are tasked with interpreting complex visual input, a sequence of questions follows: \textit{Does AI perceive illusions? Should they? And if so, what illusions are beneficial, and what should be eliminated?}

A compelling motivation comes from radiology, where human experts benefit from the \textit{Mach band illusion} \cite{mach-band}—a perceptual effect that exaggerates contrast at luminance boundaries. Radiologists unconsciously rely on this illusion to perceive subtle differences between adjacent tissues in grayscale medical images, helping to detect the low-contrast or early-stage lesions. Although this phenomenon introduces a distortion of the raw data, it improves human interpretability and diagnostic sensitivity \cite{alexander2021visual}. For AI systems tasked with similar perceptual challenges, embracing specific illusion-like mechanisms may also improve performance. Rather than eliminating all perceptual biases, it may be beneficial for AI to ``see'' like humans in domains where human vision is already optimized for critical decision-making. 

Most existing surveys classify and explain visual illusions in humans, drawing on neurobiological and perceptual theories~\cite{carbon2014understanding}. In contrast, our work explores the parallels and discrepancies between human and AI perception through the lens of visual illusions. 
We begin with an examination of how biological systems interpret visual input through context and inference. We then analyze how different classes of AI vision models—from convolutional networks to generative vision language models—respond to classic illusion stimuli. Finally, we investigate whether AI exhibits illusion-like phenomena, characterize AI-specific illusions absent in human perception, and assess their implications for human–AI perceptual alignment.

By comparing two fundamentally different ways of seeing, one biologically evolved, the other engineered, we provide the first systematic analysis of where human and AI perceptions of visual illusions converge and where uniquely artificial illusion-like behaviors arise. This comparative lens not only reveals alignment gaps and AI-specific perceptual vulnerabilities but also offers actionable insights for building vision systems that preserve human-beneficial perceptual cues while mitigating distortions that threaten trust and safety.

\section{What Are Human Visual Illusions?}

Visual illusions arise when there is a mismatch between the physical properties of a stimulus and the way it is perceived. Rather than simple perceptual errors, illusions reveal how the visual system actively interprets incomplete or ambiguous input \cite{illusion-theory}. 
Illusions can emerge at various stages of visual processing. In the first stage, retinal mechanisms enhance contrast at light-dark boundaries, leading to brightness illusions such as Mach bands \cite{mach-band}. As processing continues, orientation-sensitive neurons respond to the direction of lines and shapes, sometimes causing angular misperceptions, as in the Zöllner illusion \cite{Zöllner-Illusion-0}, where parallel lines appear to diverge or converge.
As information is integrated at higher levels, the brain combines depth, size, and motion cues, which can lead to spatial distortions like the Ponzo illusion \cite{Ponzo-Illusion}, or dynamic misperceptions such as the rotating snakes illusion \cite{Rotating-Snakes-Illusion,illusion-waving}. The brain also fills in missing information to form coherent perceptions. For example, in illusory contour illusions, people see edges or shapes that are not physically present.
Collectively, these examples underscore that perception is not a passive reflection of reality, but a product of the brain’s active interpretation, prediction, and reconstruction of visual input. While the literature presents multiple ways to classify visual illusions, this survey adopts a representative five-category framework for clarity \cite{illusionn-types}.

\subsection{Illusion Categories}

\subsubsection{Color and Brightness Illusions}

The human visual system interprets color and brightness in relative terms rather than through absolute measurements. Perception is shaped by local contrast, surrounding hues, and spatial context, enabling stability across varying illumination conditions. However, this contextual sensitivity can give rise to perceptual discrepancies, where the perceived color or brightness deviates from the physical stimulus. Figure~\ref{fig:color} presents two representative examples. In the Munker–White illusion (left), identical neutral-colored spheres appear to differ in hue due to the influence of surrounding striped patterns~\cite{Munker–White_Illusion}. In the Simultaneous Brightness Contrast illusion (right), identical gray targets appear lighter or darker depending on the luminance of their immediate background~\cite{Brightness-Contrast-Illusion}. These effects highlight the role of spatial contrast and integration mechanisms in early vision. 
Additional examples include the Cornsweet~\cite{illusionn-types}, Checker Shadow~\cite{Brightness-Contrast-Illusion}, and Watercolor illusions~\cite{illusionn-types}, all of which demonstrate how contextual modulation can override objective surface properties.

\subsubsection{Geometric-Optical Illusions}

\begin{figure}
    \centering
    \includegraphics[width=0.9\linewidth]{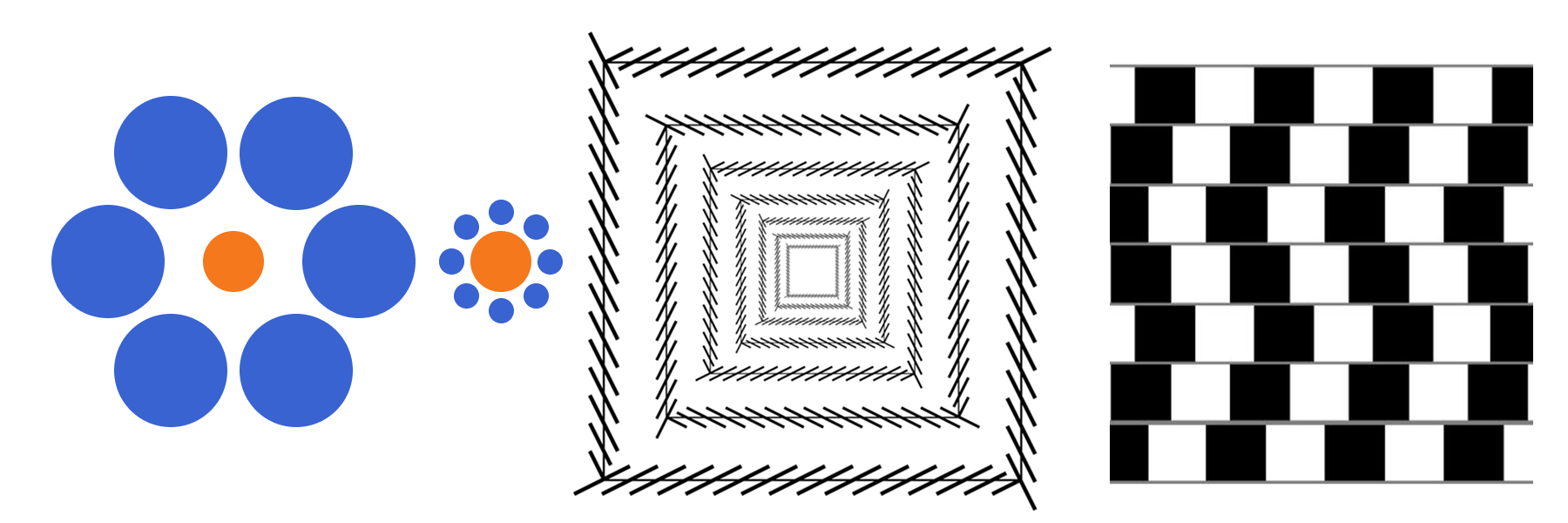}
    \caption{Geometric-Optical Illusions: Ebbinghaus Illusion (left), Zöllner Illusion (center) and Café Wall Illusion (right). Images by Akiyoshi Kitaoka.}
    \label{fig:size}
\end{figure}

Geometric-optical illusions occur when the perceived shape, size, orientation, or alignment of visual elements deviates systematically from their physical properties due to contextual influence. These illusions reveal how the visual system integrates local features with global patterns, often resulting in misperceptions that reflect the brain’s reliance on relative spatial information rather than absolute geometry. Figure~\ref{fig:size} illustrates three representative examples. In the Ebbinghaus illusion (left), two identical central circles appear different in size depending on the relative size of the surrounding disks, demonstrating context-driven size perception~\cite{Geometrical-illusion}. The Zöllner illusion (center) shows how parallel square outlines appear distorted due to intersecting diagonal lines that bias orientation judgments~\cite{Geometrical-illusion}. In the Café Wall illusion (right), horizontal gray lines appear slanted due to the alternating black and white tiles and their offset arrangement, illustrating the interaction between contrast and spatial alignment~\cite{Café-Wall-Illusion}. 
Additional geometric illusions include the Müller–Lyer illusion~\cite{Geometrical-illusion}, the Oppel–Kundt illusion~\cite{Geometrical-illusion}, and the Delboeuf illusion~\cite{Geometrical-illusion}, each highlighting distinct ways in which contextual structure shapes spatial perception.

\subsubsection{Depth and Space Illusions}

\begin{figure}
    \centering
    \includegraphics[width=0.9\linewidth]{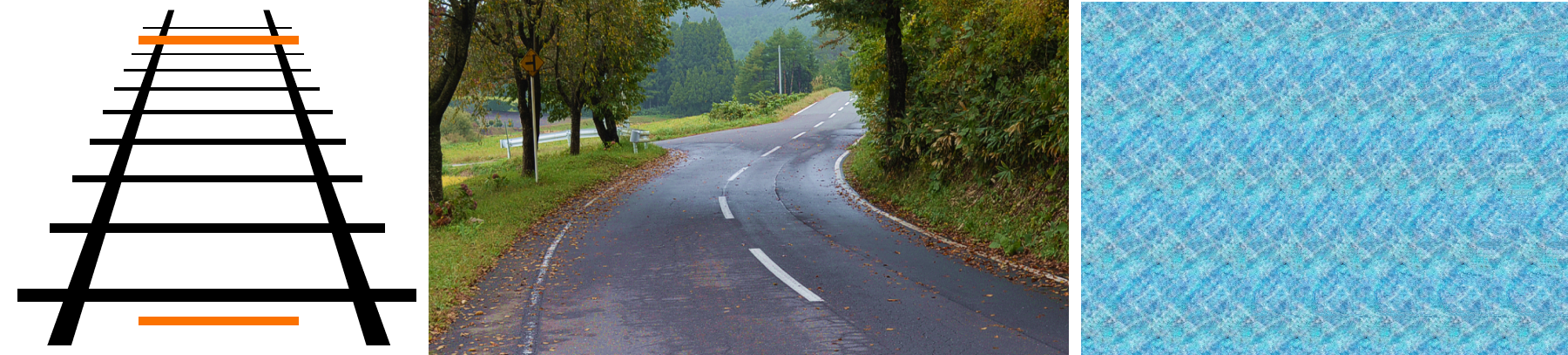}
    \caption{Depth and Space Illusions: Ponzo Illusion (left), Slope Illusion (center), and Stereograms Illusion(right). Images by Akiyoshi Kitaoka.}
    \label{fig:shape}
\end{figure}

Depth and space illusions occur when the visual system misinterprets cues related to distance, volume, or spatial configuration. These misperceptions arise because the brain infers three-dimensional structure from two-dimensional retinal images using heuristic rules such as linear perspective, shading, occlusion, and relative size. While generally effective, these strategies can produce systematic errors under certain visual conditions. Figure~\ref{fig:shape} illustrates three representative examples. In the Ponzo illusion (left), two identical orange bars placed between converging “railway” lines appear to differ in length due to the influence of linear perspective cues~\cite{Ponzo-Illusion}. The Slope illusion (center) depicts two connected slopes that both descend physically, yet one appears to ascend—an effect driven by misleading curvature and contextual features~\cite{illusion-slope}. In the Stereogram illusion (right), a three-dimensional cube emerges from a flat, repeating pattern when viewed with appropriate binocular convergence, leveraging horizontal disparities to create the impression of depth~\cite{illusion-Stereograms}. 
Additional examples of depth and space illusions include the Hering illusion~\cite{Hering-Illusion}, Autostereograms~\cite{illusion-Stereograms}, and Antigravity Hills~\cite{illusion-slope}, all of which highlight the constructive yet fallible nature of human depth perception.

\subsubsection{Motion Illusions}

\begin{figure}
    \centering
    \includegraphics[width=0.9\linewidth]{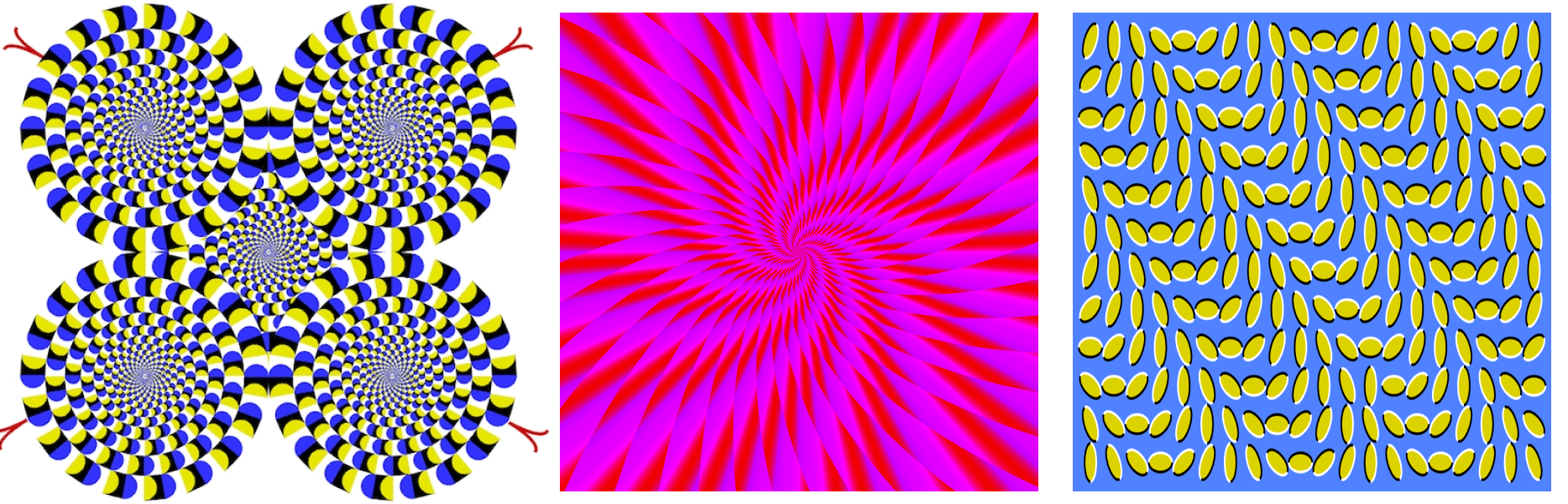}
    \caption{Motion Illusions: Rotating Snakes Illusion (left), Fraser Wilcox Illusion (center), and Waving Illusion (right). Image by Akiyoshi Kitaoka.}
    \label{fig:movement}
\end{figure}

Motion illusions arise when static images evoke a vivid perception of movement, revealing how the visual system relies on local contrast, temporal integration, and predictive mechanisms to interpret dynamic information. These illusions demonstrate the brain’s tendency to infer motion based on prior experience, spatial gradients, and micro-movements of the eyes. Figure~\ref{fig:movement} presents three illustrative examples. In the Rotating Snakes illusion (left), concentric circular patterns appear to spin continuously, an effect attributed to contrast polarity and involuntary eye movements known as microsaccades~\cite{Rotating-Snakes-Illusion}. The Fraser–Wilcox illusion (center) produces a similar swirling effect through radial gradients in luminance and color that stimulate motion-sensitive neurons in early visual areas~\cite{Rotating-Snakes-Illusion}. In the Waving illusion (right), rows of elliptical elements seem to oscillate rhythmically despite being static; this effect emerges from peripheral viewing, local luminance differences, and asymmetrical spatial arrangement, which interact with microsaccadic activity to generate perceived motion~\cite{illusion-waving}. 
Additional motion illusions such as the Peripheral Drift Illusion~\cite{illusion-waving}, Motion Aftereffect~\cite{anstis1998motion}, and Stepping Feet further illustrate~\cite{howe2006explaining} how the visual system constructs motion from spatial patterns and low-level neural dynamics.

\subsubsection{Other Cross-Domain Illusions}
\begin{figure}   % this group of images are from wikipedia
    \centering
    \includegraphics[width=0.9\linewidth]{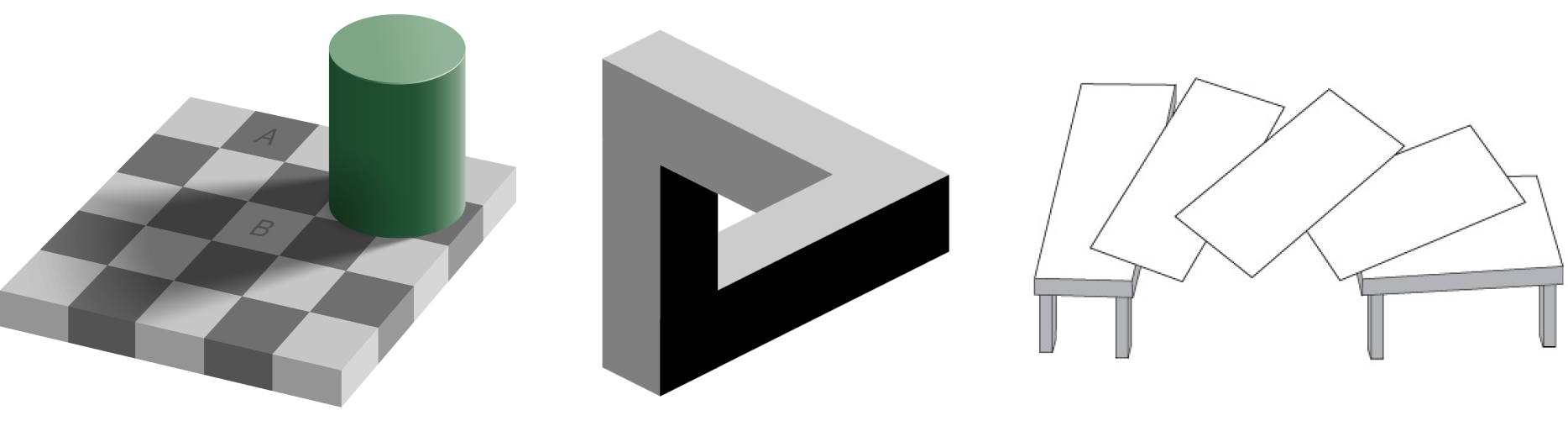}
    \caption{Other Illusions: Checker Shadow Illusion (left), Penrose Trident (center), and Shepard Tables (right). Image by Edward H. Adelson, Akiyoshi Kitaoka, and R.Shepard.}
    \label{fig:cross}
\end{figure}

Some visual illusions defy classification into a single perceptual category, as their effects emerge from interactions among multiple visual processes. These cross-domain illusions illustrate the complexity of perception, combining elements of geometry, brightness, depth, and contextual inference. Figure~\ref{fig:cross} presents three such examples, as described in The Oxford Compendium of Visual Illusions. In the Checker Shadow illusion (left),  two squares labeled A and B appear to differ in brightness despite having identical luminance. This effect arises from the interplay of shadow interpretation, 3D surface inference, and brightness constancy mechanisms. The Penrose Triangle illusion (center) presents an impossible object in which three beams form a seemingly coherent triangular structure. Although locally consistent, the global configuration defies Euclidean geometry, combining depth cues with geometric paradox. The Shepard Tables illusion features two tabletops that appear to differ in shape and size, though they are physically identical. This illusion exploits depth perspective, foreshortening, and shape constancy to produce a compelling spatial distortion. These examples highlight the richness of visual illusion~\cite{illusionn-types}.

\subsection{Are Visual Illusions Beneficial to Humans?}

Visual illusions are not mere perceptual failures but rather artifacts of the brain’s adaptive mechanisms for efficient interpretation. In many real-world scenarios, illusions serve functional purposes that enhance human performance. For instance, in medical imaging, perceptual effects like the Mach band illusion can heighten contrast sensitivity, enabling radiologists to detect subtle anatomical boundaries that might otherwise be missed \cite{medical-Mach-band}. Similarly, in traffic engineering, illusions such as converging road markings or receding arrows are deliberately employed to create a perception of increased speed, encouraging drivers to decelerate and improving safety \cite{drive}. Architects and designers also leverage perspective-based illusions to make confined spaces feel more expansive and comfortable. These applications underscore the utility of illusions as perceptual heuristics that facilitate decision-making in complex environments, even if they deviate from objective reality.

However, the same perceptual biases that support efficiency can also introduce systematic errors. In high-stakes contexts such as courtroom testimony, visual illusions can distort perception or memory—for example, misjudging distance, lighting, or motion—potentially leading to inaccurate or misleading witness statements. In advertising and marketing, visual manipulations—such as misleading scale or contrast enhancements—can distort consumer perception and influence purchasing behavior in deceptive ways. Moreover, illusions can be weaponized in digital media to construct persuasive but misleading narratives, undermining critical reasoning and public trust. These examples reveal the dual nature of visual illusions: while they are often beneficial byproducts of a predictive visual system, they can also be exploited or misfire, introducing perceptual distortions that carry real-world consequences.

\section{How Does AI Perceive Visual Information Differently from Humans?}

\begin{figure}[!t]
    \centering
    \includegraphics[width=\textwidth]{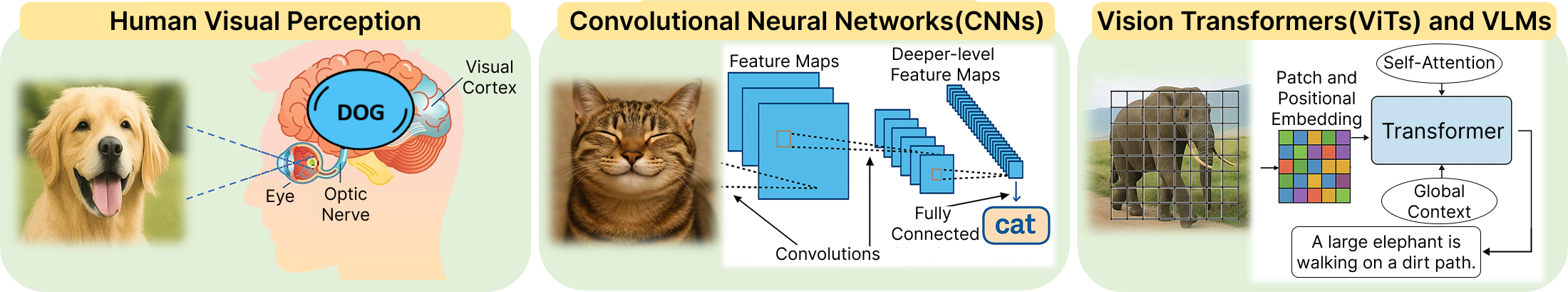}
    \caption{Contrasting visual processing paradigms. Left: \textbf{Human perception} leverages priors and semantic reasoning. Middle: \textbf{CNNs} use hierarchical feature extraction. Right: \textbf{ViTs and VLMs} enable global and multimodal understanding.}
    \label{fig:perception_comparison}
\end{figure}

\textbf{Human vision} relies on contextual inference, prior knowledge, and hierarchical processing, enabling efficient interpretation of complex scenes and ambiguity. \emph{In contrast, most deep neural networks (DNNs) process raw pixel intensities without such rich priors or adaptive mechanisms.} Classical computer vision models learn from statistical patterns in data, which limits their interpretive capacity compared to biological vision. Figure~\ref{fig:perception_comparison} summarizes the core processing pathways of human vision, convolutional neural networks (CNNs), and modern vision-language models (VLMs).

\textbf{Convolutional Neural Networks (CNNs)} \cite{krizhevsky2012imagenet} primarily at the pixel level, leveraging strong inductive biases such as spatial locality, translation invariance, and hierarchical feature composition. These biases facilitate efficient training, enable generalization from limited data, and enhance recognition performance. However, they also impose inherent architectural constraints that limit the model’s capacity to capture global context, as computations remain largely local and layer-dependent.

\textbf{Vision Transformers (ViTs)}~\cite{dosovitskiy2021imageworth16x16words} relax these constraints by eliminating traditional inductive biases. Instead of relying on convolution, ViTs divide an image into fixed-size patches and treat each as a token in a transformer architecture. Through self-attention, the model learns global relationships among all patches from the start, enabling broader contextual understanding. While more flexible and scalable, ViTs typically require more data and computation due to the absence of built-in spatial priors. Variants like Swin Transformers~\cite{liu2021swintransformerhierarchicalvision} reintroduce a hierarchical structure to balance locality and global context.

\textbf{Vision-Language Models (VLMs)} build on the ViT foundation by integrating visual and linguistic modalities. These models, such as CLIP~\cite{radford2021learningtransferablevisualmodels}, GPT-4V, and Gemini, process image patches alongside natural language input, allowing them to align visual elements with semantic context. 

This progression from pixels in CNNs to patch-based attention in ViTs to multimodal tokens in VLMs reflects an expanding field of view: from local appearance to global image structure to context-rich interpretation grounded in language. These developments mark a shift from models that detect isolated patterns to those that integrate broader context and semantic meaning.

However, despite these architectural advances, AI models still perceive the world in fundamentally different ways than humans. In ambiguous or context-dependent scenarios, human perception relies heavily on prior knowledge, experience, and expectations, while AI models operate primarily on learned statistical correlations.

The progression from pixel-based to language-informed representations has also reignited interest in identifying where and why machine perception diverges from human understanding. Humans can resolve ambiguity through context, tolerate noise or occlusion, and make high-level inferences based on minimal cues, even if that sometimes leads to perceptual illusions. In contrast, AI systems can be misled by small perturbations, irrelevant textures, or dataset biases, revealing their limited robustness and lack of common-sense reasoning. These gaps highlight the ongoing challenge of aligning machine vision with the flexibility and semantic richness of human perception.

\section{Visual Illusions in Artificial Perception}

Human and artificial perception are fundamentally different in origin, structure, and function. One is shaped by evolution and grounded in survival-driven inference; the other is engineered from data and optimized for statistical objectives. Yet a critical question remains: do AI systems perceive visual illusions the same way humans do?

\subsection{Do Computer Vision Models Have Visual Illusion?}

The first step in understanding AI perception is to ask whether computer vision (CV) models experience visual illusions in the same way humans do. Intuitively, the answer would seem to be no, given the fundamental working mechanism differences between human and artificial visual systems. Yet, surprisingly, several studies have shown that conventional image classification systems, trained only to recognize objects in natural photos, sometimes exhibit illusion-like responses that resemble human perception. For example, when shown classic illusions involving brightness or size contrast, these models produce outputs that suggest similar perceptual biases \cite{ill-1}.

Why does this happen? One explanation comes from information theory: both biological and artificial systems are optimized to extract the most informative features from limited input. As a result, they may rely on similar shortcuts, such as edge detection or contrast exaggeration, that inadvertently lead to illusions \cite{barlow-possible}. 
Another contributing factor is the nature of the data itself. Training on vast collections of real-world images can encourage models to develop internal representations that mirror the kinds of biases humans have evolved to use \cite{ill-4}. 
Finally, the design of CV systems, especially the way they compress and reconstruct visual information, may promote the emergence of these effects even without targeted instruction \cite{ill-6}.

Although human visual systems and artificial neural networks differ fundamentally in their biological and computational foundations, they often operate under comparable constraints, such as limited processing time, the need for generalization, and reliance on ambiguous or partial visual input. These shared pressures can lead both systems to adopt similar heuristic strategies, which may inadvertently give rise to illusion-like phenomena.

These findings suggest that visual illusions are not unique to human biology but can emerge from general-purpose learning systems operating under similar constraints. This convergence opens new perspectives on how perceptual biases might arise as a byproduct of efficient visual encoding—whether in brains or in AI.

\subsection{Can AI Mimic Human Illusions and Go Beyond Them?}

Researchers have also investigated whether neural networks can reproduce human perceptual responses to visual illusions by training them on illusion datasets.
% labeled datasets specifically designed for this task. 
These datasets span a range of illusion types, including brightness, size, and shape distortions, and are annotated with human judgments reflecting varying degrees of illusory effect \cite{illusion-human-data}. In a representative study, Kubota et al. \cite{ill-2} trained CNNs to match these human responses, effectively teaching the models to mimic perceptual deviations rather than physical ground truth.

Their findings reveal that, under suitable training conditions, CNNs can closely approximate human judgments across illusion types and difficulty levels. In some cases, the models not only aligned with human perception but exhibited exaggerated illusory effects, suggesting a strong internalization of contextual visual cues. This demonstrates that AI systems, when guided by perceptual annotations, can emulate the outcome of human visual processing, though the underlying mechanisms may differ fundamentally from those of biological vision.

\begin{figure}
    \centering
    \includegraphics[width=1\linewidth]{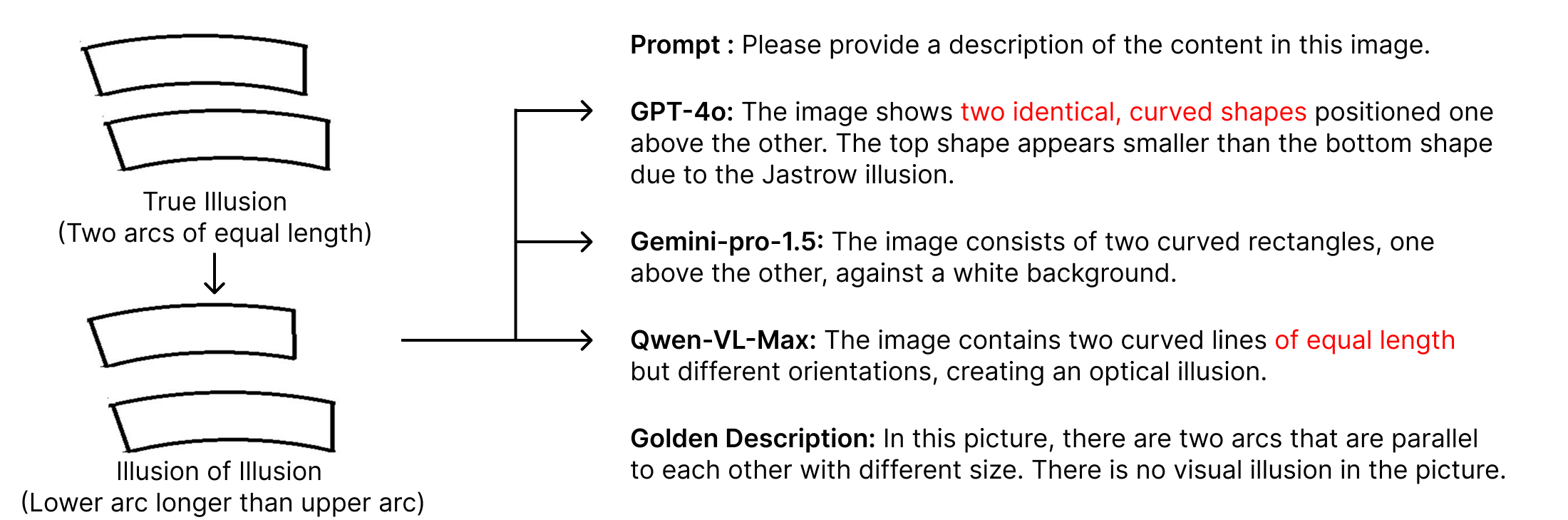}
    \caption{``Illusion of Illusion'': The lower-left arc is physically longer than the upper one, yet GPT-4o and Qwen misclassify this as a genuine illusion. Adapted from \cite{ill-13}.}
    \label{fig:fake-illusion}
\end{figure}

\subsection{AI-Specific Illusions}

AI systems not only share specific human-like perceptual errors, but also exhibit their own \textbf{distinctively artificial illusions}. These arise not from ambiguous input, but from internal limitations in model architecture, data representation, and modality alignment.

\begin{figure}
    \centering
    \includegraphics[width=1\linewidth]{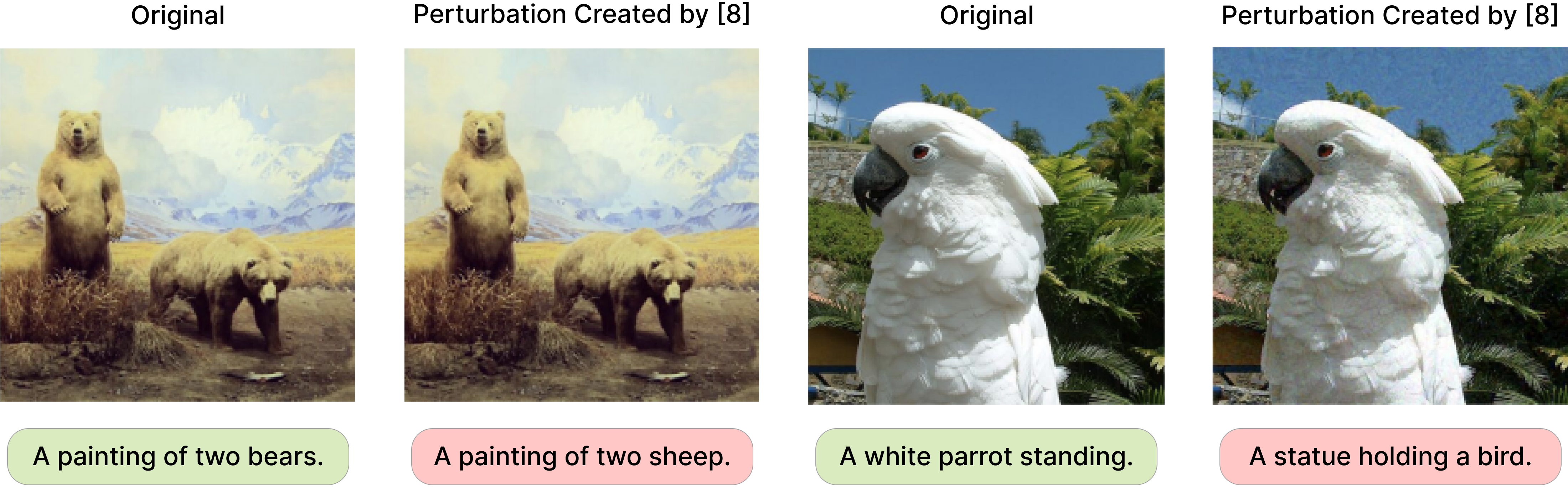}
    \caption{Pixel sensitivity in InstructBLIP~\cite{blackbox}: Tiny, invisible changes to the image lead the model to produce incorrect descriptions.}
    \vspace{-0.4cm}
    \label{fig:adversarial}
\end{figure}

A well-known example is \textit{pixel-level sensitivity}~\cite{liu2024generation}. Many vision models are easily influenced by tiny, human-invisible perturbations that shift their predictions. These vulnerabilities arise from the model's dependence on low-level statistical features rather than global semantic understanding. While humans perceive scenes holistically and tolerate noise, these models process visual input at the pixel level. As a result, adversarial examples can cause confident miscaptioning. As shown in Figure~\ref{fig:adversarial}, even minimal pixel alterations can make a captioning model describe two bears as ``two sheep in a field,'' or a parrot as a ``statue holding a bird''~\cite{blackbox}.

Another major AI illusion is \textit{hallucination} in VLMs~\cite{Hallucination}. It refers to the model generating plausible descriptions of content that does not exist in the image. These hallucinations often stem from weak visual-language grounding, where linguistic priors override perceptual input. Unlike humans, who typically signal uncertainty, these models produce confident fabrications, revealing a disconnect between what is seen and what is said.

In contrast to human illusions, which often arise from adaptive perceptual strategies, AI-specific illusions reveal the fragility of systems trained to match patterns without deeper understanding. These failures, from pixel-level attacks to confident hallucinations, are not side effects of perceptual shortcuts, but symptoms of mismatched grounding and brittle representations. Their presence challenges the assumption that AI perception aligns with human intuition. To build truly reliable systems, we must go beyond superficial success and ask not just whether models can see, but whether they understand what they see.

\subsection{Human AI Alignment with Regards to Illusion}

Alignment between human and AI perception is essential in the context of visual illusions, since perceptual mismatches can impact trust, interpretability, and safety. Generative VLMs represent a shift from traditional computer vision toward systems that combine perception, language understanding, and multimodal reasoning. Unlike conventional image classifiers, these models are capable of answering natural language questions about visual input. This makes them promising candidates for investigating whether AI systems perceive the world in ways aligned with human intuition—particularly in the context of visual illusions.

Shahgir et al.~\cite{ill-12} introduced the IllusionVQA benchmark to evaluate VLMs on illusion recognition using question-answering tasks. They found that alignment with human perception varies significantly by task: models like GPT-4 performed well on coarse localization (e.g., identifying where an illusion occurs), but struggled with fine-grained comparisons requiring subtle perceptual discrimination. Zhang et al.~\cite{ill-7} further showed that larger models tend to produce responses more consistent with human perception, suggesting that alignment improves with model scale.

A critical test of perceptual alignment involves ``illusion of illusion'' stimuli—images that visually resemble classic illusions but lack any actual illusory effect. While humans easily identify these as non-illusory, most VLMs misclassify them as genuine, suggesting that current models rely on superficial visual patterns rather than perceptual understanding (see Fig.~\ref{fig:fake-illusion}) \cite{ill-9}. Zhang et al.~\cite{ill-13} show that newer models like GPT-4o and DeepSeek-VL-7B reduce the misclassification rate to around 50\%, a notable improvement over earlier models that failed nearly completely. This progress stems not only from increased model scale but also from improved vision-language alignment, enhanced instruction-following capabilities, and more robust grounding between visual input and language. Further analysis reveals that model responses to these stimuli are highly sensitive to prompt phrasing. When prompts falsely imply the presence of an illusion (e.g., ``In this illusion...'' ), models like GPT-4o and Gemini Pro show a sharp rise in misclassifications—even for images that contain no illusion at all~\cite{ill-9}. This suggests that generative VLMs remain vulnerable to linguistic priming and texture-level biases, rather than engaging in genuine perceptual reasoning. 

These results indicate that while vision-language models are increasingly capable of matching human judgments in illusion tasks, they still lack a robust internal model of visual perception—and remain prone to language-driven biases.

\section{Conclusion}

\begin{table}
\centering
\small
\renewcommand{\arraystretch}{1.25}
\begin{tabular}{p{2.4cm} p{4.8cm} p{6.5cm}}
%\toprule
 & \textbf{Human Vision System} & \textbf{AI Vision System } \\
\midrule
Core Mechanism & Contextual inference, semantic priors, perceptual heuristics & Pixel-driven; relies on statistical regularities and task-specific training \\
\midrule
Color and Brightness Illusions & {\raggedright color and brightness through contrast and contextual cues} & 
\multirow[t]{4}{6.5cm}{\parbox[t]{6.5cm}{\footnotesize
\begin{itemize}[leftmargin=*, itemsep=1pt, topsep=0pt, parsep=0pt, partopsep=0pt, after=\vspace{-3pt}]
  \item AI can approximate human perceptual responses when trained on annotated datasets 
  \item Illusion-like behaviors may emerge even without explicit supervision 
  \item Perceptual responses vary across different model architectures and tasks
  \item Consistency in illusion-related tasks tends to increase with model scale 
  \item VLM outputs are often driven more by linguistic priors than by visual evidence
\end{itemize}}} \\
\cmidrule(lr){1-2}
Geometric-Optical Illusions & {\raggedright Interprets shape and size based on spatial arrangement and global patterns} & \\
\cmidrule(lr){1-2}
Depth and Space Illusions & Infers depth using perspective, shading, and occlusion heuristics & \\
\cmidrule(lr){1-2}
Motion Illusions & Detects motion via spatial gradients, eye movements, and prediction & \\
\addlinespace[10pt]
\midrule
Illusion of Illusion Detection & Accurately distinguishes real from meta-illusions & Often misclassifies illusion of illusions as real \\
\midrule
Pixel Sensitivity & Robust to imperceptible pixel changes due to reliance on global structure & Sensitive to small, targeted pixel perturbations \\
\midrule
Semantic Hallucination & Humans tend to hedge or withhold when unsure & VLMs often produce confident but inaccurate captions \\
\bottomrule
\end{tabular}
\caption{Comparison of human and AI vision systems in their perception of visual illusions.}

\label{tab:human-vs-ai-illusions}
\end{table}
 
Visual illusions expose both the capabilities and limitations of human and machine perception—but the underlying mechanisms are fundamentally different. Human vision is shaped by contextual reasoning, semantic knowledge, and evolutionary heuristics. In contrast, AI systems depend on statistical correlations learned from data, without embodied context or perceptual grounding. These contrasting foundations lead to a blend of overlap and divergence: AI can be trained to mimic human perceptual effects and may even develop illusion-like behaviors spontaneously. Yet it also suffers from machine-specific illusions—such as adversarial sensitivity and multimodal hallucinations—that have no human counterpart.
We summarize key differences between human and AI responses to visual illusions in Table~\ref{tab:human-vs-ai-illusions}.

Understanding these illusion-driven behaviors has important real-world implications. In medicine, illusions like the Mach band effect aid tissue boundary detection; in transportation, engineered illusions help guide driver behavior. If AI is to function safely in human environments, it must interpret such visual cues in ways consistent with human expectations. Insights from visual illusions are thus not just theoretical—they are key to designing models that see more like we do, and fail in ways we can predict, explain, and control.

\section{Acknowledgements}
We thank Sierra Liu, a high school student at Millburn High School, whose attempt to use the illusion to fool her mother, one of the authors, sparked this work; Dr. Jingyun Wang, Associate Professor at the SUNY College of Optometry, for providing background on illusions during a catch-up coffee meeting after the author was fooled and sought to understand the underlying mechanisms; Dr. Akiyoshi Kitaoka, Professor of Psychology at Ritsumeikan University, Osaka, Japan, and Dr. David Novick, Professor at the University of Texas at El Paso, for creating numerous high-quality illusion images, some of which are cited here.

\bibliographystyle{unsrtnat}
\bibliography{main}

\begin{thebibliography}{37}
\providecommand{\natexlab}[1]{#1}
\providecommand{\url}[1]{\texttt{#1}}
\expandafter\ifx\csname urlstyle\endcsname\relax
  \providecommand{\doi}[1]{doi: #1}\else
  \providecommand{\doi}{doi: \begingroup \urlstyle{rm}\Url}\fi

\bibitem[Howe(2005)]{Munker–White_Illusion}
Piers D~L Howe.
\newblock White's effect: Removing the junctions but preserving the strength of the illusion.
\newblock \emph{Perception}, 34\penalty0 (5):\penalty0 557--564, 2005.
\newblock \doi{10.1068/p5414}.
\newblock URL \url{https://doi.org/10.1068/p5414}.
\newblock PMID: 15991692.

\bibitem[Kingdom(2014)]{mach-band}
Frederick~AA Kingdom.
\newblock Mach bands explained by response normalization.
\newblock \emph{Frontiers in human neuroscience}, 8:\penalty0 843, 2014.

\bibitem[Alexander et~al.(2021{\natexlab{a}})Alexander, Yazdanie, Waite, Chaudhry, Kolla, Macknik, and Martinez-Conde]{alexander2021visual}
Robert~G Alexander, Fahd Yazdanie, Stephen Waite, Zeshan~A Chaudhry, Srinivas Kolla, Stephen~L Macknik, and Susana Martinez-Conde.
\newblock Visual illusions in radiology: untrue perceptions in medical images and their implications for diagnostic accuracy.
\newblock \emph{Frontiers in Neuroscience}, 15:\penalty0 629469, 2021{\natexlab{a}}.

\bibitem[Carbon(2014)]{carbon2014understanding}
Claus-Christian Carbon.
\newblock Understanding human perception by human-made illusions.
\newblock \emph{Frontiers in human neuroscience}, 8:\penalty0 566, 2014.

\bibitem[Tyler(2022)]{illusion-theory}
Christopher~W Tyler.
\newblock The nature of illusions: A new synthesis based on verifiability.
\newblock \emph{Frontiers in Human Neuroscience}, 16:\penalty0 875829, 2022.

\bibitem[Kitaoka and Ishihara(2000)]{Zöllner-Illusion-0}
Aktyoshi Kitaoka and Masami Ishihara.
\newblock Three elemental illusions determine thez{\"o}llner illusion.
\newblock \emph{Perception \& psychophysics}, 62:\penalty0 569--575, 2000.

\bibitem[Yildiz et~al.(2022)Yildiz, Sperandio, Kettle, and Chouinard]{Ponzo-Illusion}
Gizem~Y Yildiz, Irene Sperandio, Christine Kettle, and Philippe~A Chouinard.
\newblock A review on various explanations of ponzo-like illusions.
\newblock \emph{Psychonomic Bulletin \& Review}, pages 1--28, 2022.

\bibitem[Conway et~al.(2005)Conway, Kitaoka, Yazdanbakhsh, Pack, and Livingstone]{Rotating-Snakes-Illusion}
Bevil~R. Conway, Akiyoshi Kitaoka, Arash Yazdanbakhsh, Christopher~C. Pack, and Margaret~S. Livingstone.
\newblock Neural basis for a powerful static motion illusion.
\newblock \emph{Journal of Neuroscience}, 25\penalty0 (23):\penalty0 5651--5656, 2005.
\newblock ISSN 0270-6474.
\newblock \doi{10.1523/JNEUROSCI.1084-05.2005}.
\newblock URL \url{https://www.jneurosci.org/content/25/23/5651}.

\bibitem[Faubert and Herbert(1999)]{illusion-waving}
Jocelyn Faubert and Andrew~M Herbert.
\newblock The peripheral drift illusion: A motion illusion in the visual periphery.
\newblock \emph{Perception}, 28\penalty0 (5):\penalty0 617--621, 1999.
\newblock \doi{10.1068/p2825}.
\newblock URL \url{https://doi.org/10.1068/p2825}.
\newblock PMID: 10664757.

\bibitem[Shapiro and Todorovic(2016)]{illusionn-types}
Arthur~G Shapiro and Dejan Todorovic.
\newblock \emph{The Oxford compendium of visual illusions}.
\newblock Oxford University Press, 2016.

\bibitem[Williams et~al.(1998)Williams, McCoy, and Purves]{Brightness-Contrast-Illusion}
S~Mark Williams, Allison~N McCoy, and Dale Purves.
\newblock The influence of depicted illumination on brightness.
\newblock \emph{Proceedings of the National Academy of Sciences}, 95\penalty0 (22):\penalty0 13296--13300, 1998.

\bibitem[Ninio(2014)]{Geometrical-illusion}
Jacques Ninio.
\newblock Geometrical illusions are not always where you think they are: a review of some classical and less classical illusions, and ways to describe them.
\newblock \emph{Frontiers in human neuroscience}, 8:\penalty0 856, 2014.

\bibitem[Gregory and Heard(1979)]{Café-Wall-Illusion}
Richard~L Gregory and Priscilla Heard.
\newblock Border locking and the café wall illusion.
\newblock \emph{Perception}, 8\penalty0 (4):\penalty0 365--380, 1979.
\newblock \doi{10.1068/p080365}.
\newblock URL \url{https://doi.org/10.1068/p080365}.
\newblock PMID: 503767.

\bibitem[Bressan et~al.(2003)Bressan, Garlaschelli, and Barracano]{illusion-slope}
Paola Bressan, Luigi Garlaschelli, and Monica Barracano.
\newblock Antigravity hills are visual illusions.
\newblock \emph{Psychological Science}, 14\penalty0 (5):\penalty0 441--449, 2003.
\newblock \doi{10.1111/1467-9280.02451}.
\newblock URL \url{https://doi.org/10.1111/1467-9280.02451}.
\newblock PMID: 12930474.

\bibitem[Yankelevsky et~al.(2016)Yankelevsky, Shvartz, Avraham, and Bruckstein]{illusion-Stereograms}
Yael Yankelevsky, Ishai Shvartz, Tamar Avraham, and Alfred~M. Bruckstein.
\newblock Depth perception in autostereograms: 1/f noise is best.
\newblock \emph{J. Opt. Soc. Am. A}, 33\penalty0 (2):\penalty0 149--159, Feb 2016.
\newblock \doi{10.1364/JOSAA.33.000149}.
\newblock URL \url{https://opg.optica.org/josaa/abstract.cfm?URI=josaa-33-2-149}.

\bibitem[Howe and Purves(2005)]{Hering-Illusion}
Catherine~Q Howe and Dale Purves.
\newblock Natural-scene geometry predicts the perception of angles and line orientation.
\newblock \emph{Proceedings of the National Academy of Sciences}, 102\penalty0 (4):\penalty0 1228--1233, 2005.

\bibitem[Anstis et~al.(1998)Anstis, Verstraten, and Mather]{anstis1998motion}
Stuart Anstis, Frans~AJ Verstraten, and George Mather.
\newblock The motion aftereffect.
\newblock \emph{Trends in cognitive sciences}, 2\penalty0 (3):\penalty0 111--117, 1998.

\bibitem[Howe et~al.(2006)Howe, Thompson, Anstis, Sagreiya, and Livingstone]{howe2006explaining}
Piers~DL Howe, Peter~G Thompson, Stuart~M Anstis, Hersh Sagreiya, and Margaret~S Livingstone.
\newblock Explaining the footsteps, belly dancer, wenceslas, and kickback illusions.
\newblock \emph{Journal of vision}, 6\penalty0 (12):\penalty0 5--5, 2006.

\bibitem[Alexander et~al.(2021{\natexlab{b}})Alexander, Yazdanie, Waite, Chaudhry, Kolla, Macknik, and Martinez-Conde]{medical-Mach-band}
Robert~G Alexander, Fahd Yazdanie, Stephen Waite, Zeshan~A Chaudhry, Srinivas Kolla, Stephen~L Macknik, and Susana Martinez-Conde.
\newblock Visual illusions in radiology: untrue perceptions in medical images and their implications for diagnostic accuracy.
\newblock \emph{Frontiers in Neuroscience}, 15:\penalty0 629469, 2021{\natexlab{b}}.

\bibitem[Garach et~al.(2022)Garach, Calvo, and {De Oña}]{drive}
Laura Garach, Francisco Calvo, and Juan {De Oña}.
\newblock The effect of widening longitudinal road markings on driving speed perception.
\newblock \emph{Transportation Research Part F: Traffic Psychology and Behaviour}, 88:\penalty0 141--154, 2022.
\newblock ISSN 1369-8478.
\newblock \doi{https://doi.org/10.1016/j.trf.2022.05.021}.
\newblock URL \url{https://www.sciencedirect.com/science/article/pii/S1369847822001164}.

\bibitem[Krizhevsky et~al.(2012)Krizhevsky, Sutskever, and Hinton]{krizhevsky2012imagenet}
Alex Krizhevsky, Ilya Sutskever, and Geoffrey~E Hinton.
\newblock Imagenet classification with deep convolutional neural networks.
\newblock In \emph{Advances in neural information processing systems 25}, 2012.

\bibitem[Dosovitskiy et~al.(2021)Dosovitskiy, Beyer, Kolesnikov, Weissenborn, Zhai, Unterthiner, Dehghani, Minderer, Heigold, Gelly, Uszkoreit, and Houlsby]{dosovitskiy2021imageworth16x16words}
Alexey Dosovitskiy, Lucas Beyer, Alexander Kolesnikov, Dirk Weissenborn, Xiaohua Zhai, Thomas Unterthiner, Mostafa Dehghani, Matthias Minderer, Georg Heigold, Sylvain Gelly, Jakob Uszkoreit, and Neil Houlsby.
\newblock An image is worth 16x16 words: Transformers for image recognition at scale.
\newblock In \emph{International Conference on Learning Representations (ICLR)}, 2021.
\newblock URL \url{https://openreview.net/forum?id=YicbFdNTTy}.

\bibitem[Liu et~al.(2021)Liu, Lin, Cao, Hu, Wei, Zhang, Lin, and Guo]{liu2021swintransformerhierarchicalvision}
Ze~Liu, Yutong Lin, Yue Cao, Han Hu, Yixuan Wei, Zheng Zhang, Stephen Lin, and Baining Guo.
\newblock Swin transformer: Hierarchical vision transformer using shifted windows.
\newblock In \emph{2021 IEEE/CVF International Conference on Computer Vision (ICCV)}, pages 9992--10002, 2021.
\newblock \doi{10.1109/ICCV48922.2021.00986}.

\bibitem[Radford et~al.(2021)Radford, Kim, Hallacy, Ramesh, Goh, Agarwal, Sastry, Askell, Mishkin, Clark, Krueger, and Sutskever]{radford2021learningtransferablevisualmodels}
Alec Radford, Jong~Wook Kim, Chris Hallacy, Aditya Ramesh, Gabriel Goh, Sandhini Agarwal, Girish Sastry, Amanda Askell, Pamela Mishkin, Jack Clark, Gretchen Krueger, and Ilya Sutskever.
\newblock Learning transferable visual models from natural language supervision.
\newblock In \emph{Proceedings of the 38th International Conference on Machine Learning (ICML)}, volume 139 of \emph{Proceedings of Machine Learning Research}, pages 8748--8763. PMLR, 2021.

\bibitem[Benjamin et~al.(2019)Benjamin, Qiu, Zhang, Kording, and Stocker]{ill-1}
Ari Benjamin, Cheng Qiu, Ling-Qi Zhang, Konrad Kording, and Alan Stocker.
\newblock Shared visual illusions between humans and artificial neural networks.
\newblock In \emph{2019 Conference on Cognitive Computational Neuroscience}, volume~10, pages 2019--1299. Cognitive Computational Neuroscience, 2019.

\bibitem[Barlow et~al.(1961)]{barlow-possible}
Horace~B Barlow et~al.
\newblock Possible principles underlying the transformation of sensory messages.
\newblock \emph{Sensory communication}, 1\penalty0 (01):\penalty0 217--233, 1961.

\bibitem[Flachot et~al.(2022)Flachot, Akbarinia, Schütt, Fleming, Wichmann, and Gegenfurtner]{ill-4}
Alban Flachot, Arash Akbarinia, Heiko~H. Schütt, Roland~W. Fleming, Felix~A. Wichmann, and Karl~R. Gegenfurtner.
\newblock Deep neural models for color classification and color constancy.
\newblock \emph{Journal of Vision}, 22\penalty0 (4):\penalty0 17--17, 03 2022.
\newblock ISSN 1534-7362.
\newblock \doi{10.1167/jov.22.4.17}.
\newblock URL \url{https://doi.org/10.1167/jov.22.4.17}.

\bibitem[Mukherjee et~al.(2024)Mukherjee, Paul, and Ghosh]{ill-6}
Amrita Mukherjee, Avijit Paul, and Kuntal Ghosh.
\newblock Deep learning models for perception of brightness related illusions.
\newblock \emph{Applied Intelligence}, 54\penalty0 (21):\penalty0 10259--10283, 2024.

\bibitem[Fermüller et~al.(2010)Fermüller, Ji, and Kitaoka]{illusion-human-data}
Cornelia Fermüller, Hui Ji, and Akiyoshi Kitaoka.
\newblock Illusory motion due to causal time filtering.
\newblock \emph{Vision Research}, 50\penalty0 (3):\penalty0 315--329, 2010.
\newblock ISSN 0042-6989.
\newblock \doi{https://doi.org/10.1016/j.visres.2009.11.021}.
\newblock URL \url{https://www.sciencedirect.com/science/article/pii/S0042698909005331}.

\bibitem[Kubota et~al.(2021)Kubota, Hiyama, and Inami]{ill-2}
Yuki Kubota, Atsushi Hiyama, and Masahiko Inami.
\newblock A machine learning model perceiving brightness optical illusions: Quantitative evaluation with psychophysical data.
\newblock In \emph{Proceedings of the Augmented Humans International Conference 2021}, AHs '21, page 174–182, New York, NY, USA, 2021. Association for Computing Machinery.
\newblock ISBN 9781450384285.
\newblock \doi{10.1145/3458709.3458952}.
\newblock URL \url{https://doi.org/10.1145/3458709.3458952}.

\bibitem[Zhang et~al.(2025)Zhang, Zhang, Wei, Liu, Zhai, and Min]{ill-13}
Yiming Zhang, Zicheng Zhang, Xinyi Wei, Xiaohong Liu, Guangtao Zhai, and Xiongkuo Min.
\newblock Illusionbench: A large-scale and comprehensive benchmark for visual illusion understanding in vision-language models, 2025.
\newblock URL \url{https://arxiv.org/abs/2501.00848}.

\bibitem[Liu et~al.(2024{\natexlab{a}})Liu, Li, Guo, Liu, Tang, and Nie]{liu2024generation}
Jiangfan Liu, Yishan Li, Yanming Guo, Yu~Liu, Jun Tang, and Ying Nie.
\newblock Generation and countermeasures of adversarial examples on vision: a survey.
\newblock \emph{Artificial Intelligence Review}, 57\penalty0 (8):\penalty0 199, 2024{\natexlab{a}}.

\bibitem[Cheng et~al.(2024)Cheng, Miao, Dong, Yang, Gao, and Zhu]{blackbox}
Shuyu Cheng, Yibo Miao, Yinpeng Dong, Xiao Yang, Xiao-Shan Gao, and Jun Zhu.
\newblock Efficient black-box adversarial attacks via bayesian optimization guided by a function prior.
\newblock In \emph{Proceedings of the 41st International Conference on Machine Learning}, ICML'24. JMLR.org, 2024.

\bibitem[Liu et~al.(2024{\natexlab{b}})Liu, Xue, Chen, Chen, Zhao, Wang, Hou, Li, and Peng]{Hallucination}
Hanchao Liu, Wenyuan Xue, Yifei Chen, Dapeng Chen, Xiutian Zhao, Ke~Wang, Liping Hou, Rongjun Li, and Wei Peng.
\newblock A survey on hallucination in large vision-language models, 2024{\natexlab{b}}.
\newblock URL \url{https://arxiv.org/abs/2402.00253}.

\bibitem[Shahgir et~al.(2024)Shahgir, Sayeed, Bhattacharjee, Ahmad, Dong, and Shahriyar]{ill-12}
Haz~Sameen Shahgir, Khondker~Salman Sayeed, Abhik Bhattacharjee, Wasi~Uddin Ahmad, Yue Dong, and Rifat Shahriyar.
\newblock Illusion{VQA}: A challenging optical illusion dataset for vision language models.
\newblock In \emph{First Conference on Language Modeling}, 2024.
\newblock URL \url{https://openreview.net/forum?id=7ysaJGs7zY}.

\bibitem[Zhang et~al.(2023)Zhang, Pan, Zhou, Pan, and Chai]{ill-7}
Yichi Zhang, Jiayi Pan, Yuchen Zhou, Rui Pan, and Joyce Chai.
\newblock Grounding visual illusions in language: Do vision-language models perceive illusions like humans?
\newblock In Houda Bouamor, Juan Pino, and Kalika Bali, editors, \emph{Proceedings of the 2023 Conference on Empirical Methods in Natural Language Processing}, pages 5718--5728, Singapore, December 2023. Association for Computational Linguistics.
\newblock \doi{10.18653/v1/2023.emnlp-main.348}.
\newblock URL \url{https://aclanthology.org/2023.emnlp-main.348/}.

\bibitem[Ullman(2024)]{ill-9}
Tomer Ullman.
\newblock The illusion-illusion: Vision language models see illusions where there are none, 2024.
\newblock URL \url{https://arxiv.org/abs/2412.18613}.

\end{thebibliography}

\end{document}